\documentclass{article} 
\usepackage{iclr2015,times}
\usepackage{hyperref}
\usepackage{url}
\usepackage{graphicx}

\title{Purine: A bi-graph based deep learning framework}

\author{
Min Lin\textsuperscript{1,2}, Shuo Li\textsuperscript{3}, Xuan
Luo\textsuperscript{3} \& Shuicheng Yan\textsuperscript{2} \\
1. Graduate School of integrated Sciences and Engineering \\
2. Department of Electrical and Computer Engineering \\
National University of Singapore \\
3. Zhiyuan College, Shanghai Jiao Tong University \\
\texttt{\{linmin, eleyans\}@nus.edu.sg} \\
\texttt{li3shuo1@gmail.com} \\
\texttt{roxanneluo@sjtu.edu.cn} \\
}

\iclrfinalcopy 

\begin{document}

\maketitle

\begin{abstract}
  In this paper, we introduce a novel deep learning framework, termed
  Purine. In Purine, a deep network is expressed as a bipartite graph
  (bi-graph), which is composed of interconnected operators and data
  tensors. With the bi-graph abstraction, networks are easily solvable
  with event-driven task dispatcher. We then demonstrate that
  different parallelism schemes over GPUs and/or CPUs on single or
  multiple PCs can be universally implemented by graph
  composition. This eases researchers from coding for various
  parallelization schemes, and the same dispatcher can be used for
  solving variant graphs. Scheduled by the task dispatcher, memory
  transfers are fully overlapped with other computations, which
  greatly reduces the communication overhead and helps us achieve
  approximate linear acceleration.
\end{abstract}

\section{Introduction}
The need for training deep neural networks on large-scale datasets has
motivated serveral research works that aim to accelerate the training
process by parallelising the training on multiple CPUs or GPUs. There
are two different ways to parallelize the training. (1) Model
parallelism: the model is distributed to different computing nodes
 \citep{sutskever2014sequence} (2) Data parallelism: different nodes
train on different samples for the same
model \citep{seide20141,chilimbi2014project}. Some of the works even
use a hybrid of
them \citep{krizhevsky2014one,dean2012large,le2013building}.  For data
parallelism, there are also two schemes regarding communication
between the peers. (1) the allreduce approach where all updates from
the peers are aggregated at the synchronization point and the averaged
update is broadcasted back to the
peers \citep{seide20141,krizhevsky2014one}. (2) the parameter server
approach handles the reads and writes of the parameters
asynchronously \citep{dean2012large,le2013building,chilimbi2014project}.
Efficient implementations of the various parallelization schemes
described by previous works are non-trivial.

To facilitate the implementation of various parallelization schemes,
we built a bigraph-based deep learning framework called ``Purine''. It
is named ``Purine'', which is an analog of caffeine in molecular
structure, because we benefited a lot from the open source Caffe
framework \citep{jia2014caffe} in our research and the math functions
used in Purine are ported from Caffe.

\section{Bi-Graph Abstraction}
\begin{figure}[h!]
    \centering
    \includegraphics[width=0.8\textwidth]{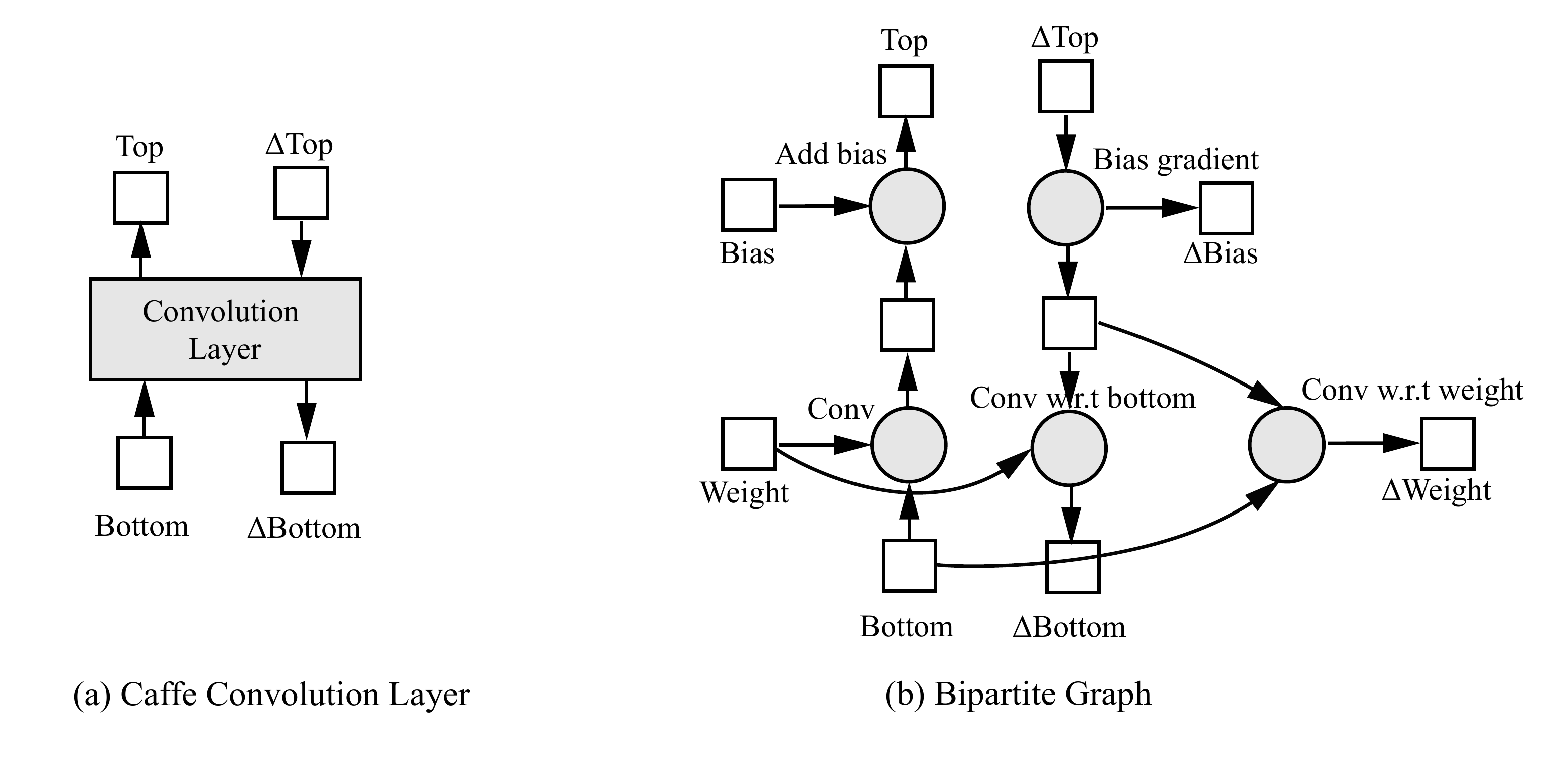}
    \caption{(a) shows the convolution layer defined in Caffe together
      with its inputs and outputs. (b) is the corresponding bipartite
      graph that describes the underlying computation inside the
      convolution layer. There are two types of vertices in the
      Bi-Graph.  Boxes represent data tensors and the circles
      represent operators. Operators are functions of the incoming
      tensors and the results of the functions are put in the outgoing
      tensors.}
    \label{fig:conv_dataflow_graph}
\end{figure}

Purine abstracts the processing procedure of deep neural networks into
directed bipartite graphs (Bi-Graphs). The Bi-Graph contains two types
of vertices, tensors and operators. Directed edges are only between
tensors and operators and there is no interconnection within tensors or
operators. Figure \ref{fig:conv_dataflow_graph} illustrates the
Bi-Graph for the convolution layer defined in Caffe.

All feed-forward neural nets can be represented by a directed acyclic
bipartite graph, which can be solved by a universal task dispatcher.
There are several works that use similar abstractions. For example,
the dataflow graph in Dryad \citep{isard2007dryad} and Pig Latin
\citep{olston2008pig} are the same as the Bi-Graph abstraction
introduced in this paper. Graphlab \citep{low2010graphlab} proposed a
more general abstraction which is applicable to iterative
algorithms. However, these systems are designed for general problems
and do not support GPU. Theano \citep{bergstra+al:2010-scipy} compiles
math expressions and their symbolic differentiations into graphs for
evalutation. Though it supports GPU and is widely used for deep
learning, the ability to parallelize over multiple GPUs and over GPUs on
different machines is not complete.

\subsection{Task Dispatcher}
Purine solves the Bi-Graph by scheduling the operators within the
Bi-Graph with an event-driven task dispatcher. Execution of an operator
is triggered when all the incoming tensors are ready. A tensor is
ready when all its incoming operators have completed computation. The
computation of the Bi-Graph starts from the sources of the graph
and stops when all the sinks are reached. This process is scheduled by
the task dispatcher.

\subsection{Iterations}
\begin{figure}[h!]
    \centering
    \includegraphics[width=0.8\textwidth]{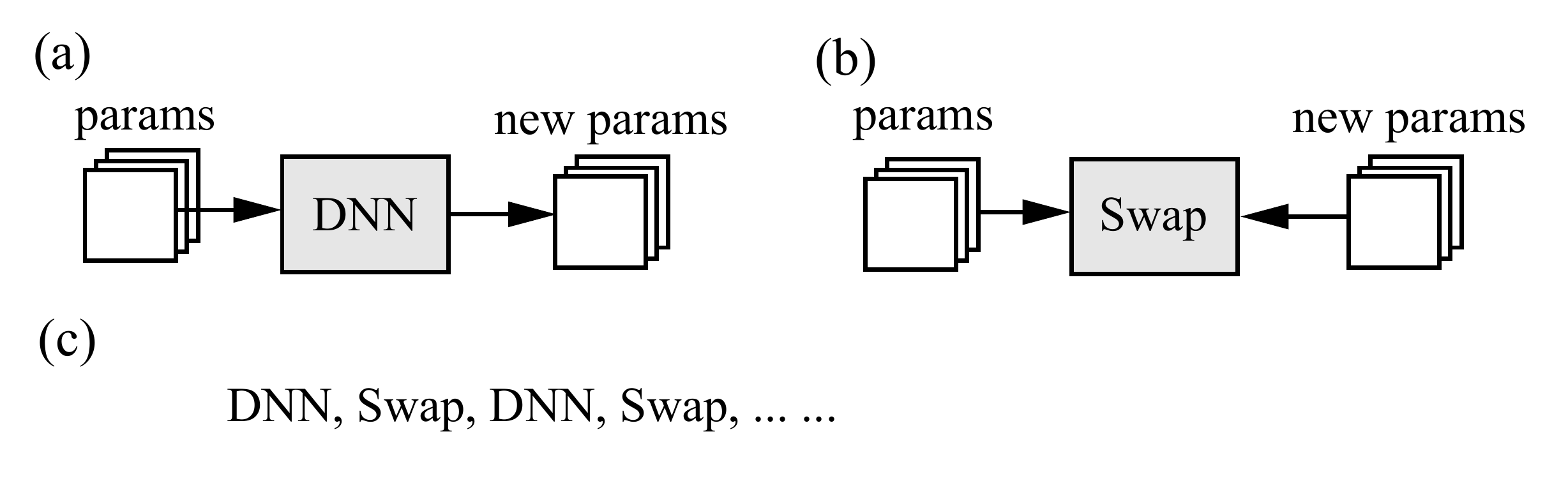}
    \caption{Implementation of SGD by Graph iteration. Every iteration
      of SGD calculates a modification of the network parameter, and
      updates the parameter before the next iteration. Since direct
      updating the network parameter would form a cyclic loop in the
      graph, it is dissected into two parts. (a) The DNN graph
      calculates the updated parameters and places them in ``new
      params'', (b) The swap graph will swap the memory address of the
      ``new'' and ``old'' parameters. As a whole, SGD is implemented
      by iterating the two graphs as in (c).}
    \label{fig:iterations}
\end{figure}

Although it has been argued in \citep{low2010graphlab} that the
directed acyclic graph could not effectively express iterative
algorithms as the graph structure would depend on the number of
iterations. We overcome this by iteration of the graphs. Because the
task dispatcher waits until all the sinks of the graph are reached, it
acts as a synchronization point. Thus parallelizable operations can be
put in a single graph, while sequential tasks (iterations) are
implemented by iteration of graphs. A concrete example is shown in
Figure \ref{fig:iterations}.

\section{Parallelization}
Parallelization of the Bi-Graph on a cluster of CPUs or GPUs or mixed
can be easily implemented by introducing a ``location'' property for
the tensors and operators. The ``location'' property uniquely
identifies the computation resource (CPUs/GPUs) on which a
tensor/operator should be allocated. The ``location'' property
comprises two fields: hostname and device id. In a multi-machine
cluster, hostname identifies the machine that the vertice resides
on. Device id specifies whether the tensor/operator should be
allocated on CPU or GPU and the ordinal of the GPU if there are
multiple GPUs installed on a single machine. Besides the ``location''
property, another property ``thread'' is assigned to operators because
both CPU and GPU support multithreading. Operators with the same
thread id will be queued in the same thread, while those with
different ids are parallelized whenever possible. It is up to the user
to decide the assignment of the graph over the computation resources.

\subsection{Copy Operator}
In the multidevice setting, data located on one device are not
directly accessible by operators on another. Thus a special ``Copy''
operator is introduced to cross the boundary, connecting parts of
the Bi-Graph on individual devices. The Copy operators, just like other
operators, are scheduled by the task dispatcher. Therefore it is
straightforward to overlap copy operations with other computation tasks
by assigning different threads to them.

\subsection{Task Dispatcher}
In the case of single machine and multiple devices, only one
dispatcher process is launched. Operators are associated to their
threads and scheduled by the global task dispatcher.  In the case of multiple
machines and multiple devices, individual dispatcher processes are
launched on each of the machines. Copy operators that copy data from
machine A to machine B are sinks on machine A and sources on machine
B. This way, each machine only needs to schedule its own subgraph and
no global scheduling mechanism or communication between dispatchers is
necessary.

\subsection{Model Parallelism}
\begin{figure}[h!]
    \centering
    \includegraphics[width=0.8\textwidth]{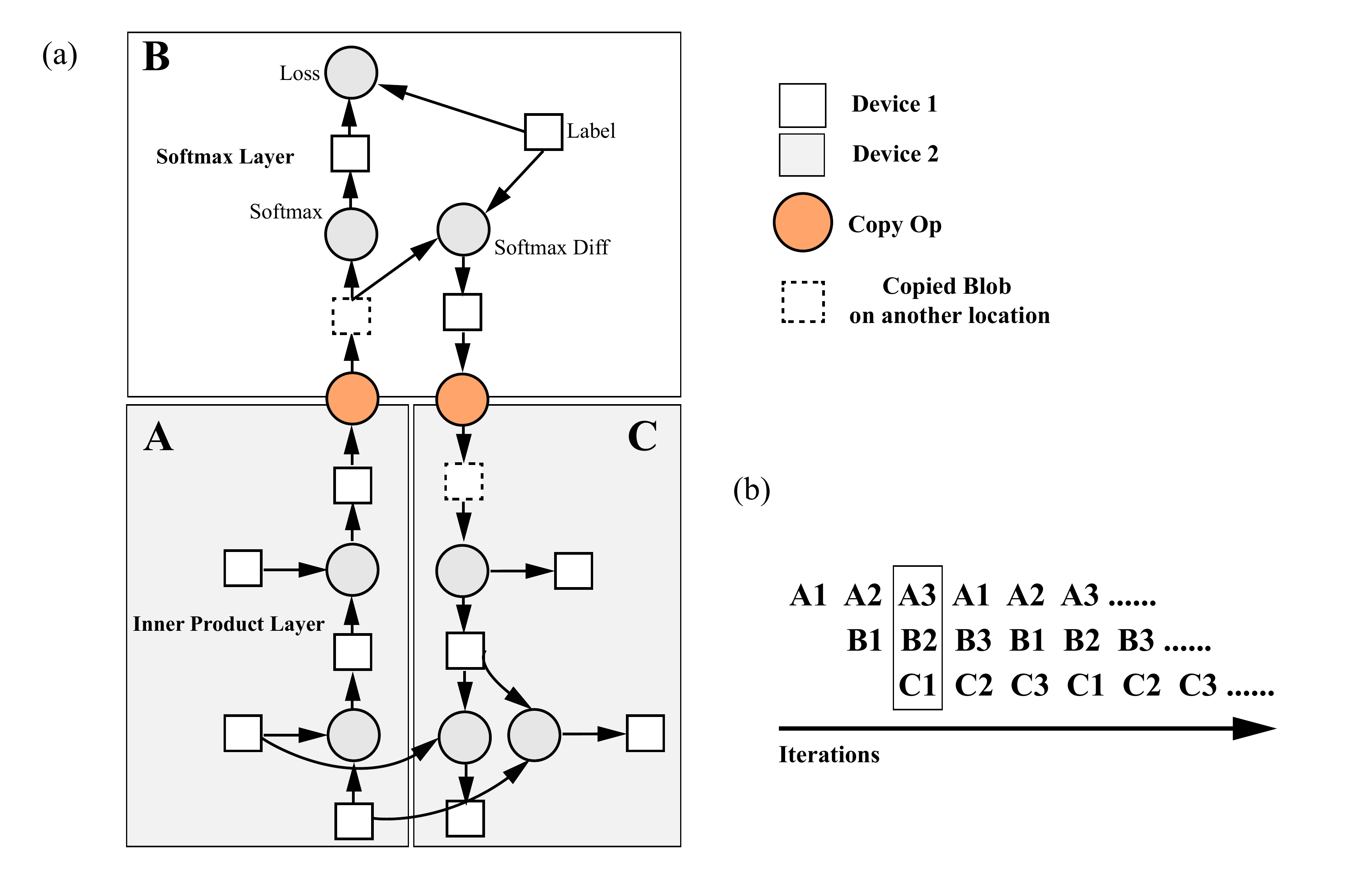}
    \caption{Implementing model parallelism in Purine. (a) The
      two-layer network can be divided into three subgraphs which
      execute in sequence. (b) The network is replicated three times
      and executed in order.}
    \label{fig:pipeline}
\end{figure}

We demonstrate how model parallelism can be implemented in Purine
by taking a two-layer fully connected neural network as example. It
can be extended to deeper networks easily. As is shown in Figure
\ref{fig:pipeline}, execution of the two-layer network can be divided
into three sequential steps. They are labeled as A, B, C
correspondingly. To keep resources busy all the time, the network is
replicated three times and executed in order.

\subsection{Data Parallelism}
Data parallelism is a simple yet straightforward way to parallelize
deep networks. In data parallelism, computation peers each keep a
replicate of the deep network. The communication between peers can be
either synchronous or asynchronous. In the synchonous case, the
gradients from peers are gathered by the parameter server. The updated
parameter is calculated and copied back to all the peers.

A hybrid of data parallelism and model parallelism has previously been
proposed by \citet{krizhevsky2014one} in which the convolution layers
use data parallelism and fully connected layers use model
parallelism. This is based on the observation that the number of
parameters is large and thus the communication cost is big for fully
connected layers. The hybrid approach greatly reduces the
communication cost. A different approach to reduce communication
overhead is to overlap the data transfer with computations.
Double buffering is proposed by \citet{seide20141} to break a minibatch
in half and exchange the gradients of the first half while doing
computaion of the second half.

\begin{figure}[h!]
    \centering
    \includegraphics[width=0.6\textwidth]{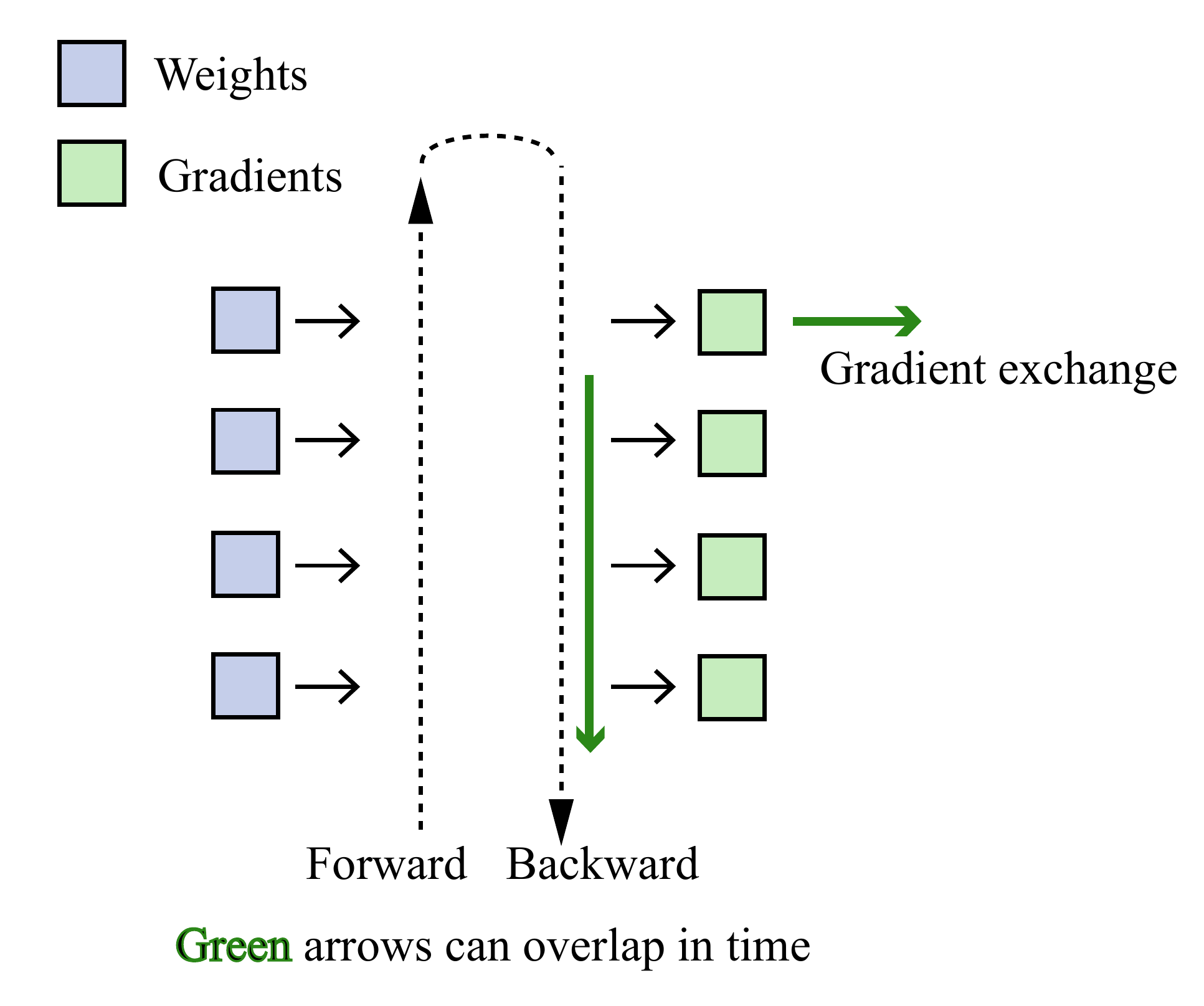}
    \caption{Overlapping communication with computation.}
    \label{fig:overlap}
\end{figure}

With the scheduling of the task dispatcher in Purine, we propose a
more straightforward way to hide the communication overhead. We show
that data parallelism is feasible even for fully connected
layers, especially when the network is very deep. Since the fully
connected layers are usually at the top of the neural networks,
exchange of the parameter gradients can be overlapped with the
backward computation of the convolution layers. As is shown in Figure
\ref{fig:overlap}, exchange of gradients in the higher layer can be
overlapped with the computation of lower layers. Gradient exchange of
lower layers could be less of a problem because they usually have a
much smaller number of parameters.

\section{Results}
We carried out experiments on the Purine framework with data
parallelism on GoogLeNet \citep{szegedy2014going}.  Data parallelism
often results in larger batch sizes, which are unfavorable for SGD
convergence demonstrated by previous studies. In this paper we ignored
the possible change in convergence rate but instead studied how much
more data can be processed per unit time with the parallelization.

We compared the number of images processed per second for GoogLeNet
with different numbers of GPUs for data parallelism. The batch size we
use is 128 per GPU. There are 3 GPUs installed on each workstation,
interconnected with 10 Gigabit Ethernet.

As is shown in Table \ref{table:persecond}, the speed increases
linearly with more GPUs added. The speed is faster than the previous
version of this paper because we upgraded the CUDNN library to version
2, which is faster compared to version 1.

Note that the machines are connected by 10 gigabit ethernet and thus data
on GPU need to go through CPU memory to be tranferred over the
ethernet. Even with this limitation, the speed up is linear thanks to
the overlapping of communication with computation.

\begin{table}[h]
\caption{Number of images per second with increasing number of
  GPUs. (GPU number smaller than or equal to 3 are tested on single
  machine. Performances of GPU number larger than 3 are on different
  machines interconnected with 10 Gigabit Ethernet.)}
\label{table:persecond}
\begin{center}
\begin{tabular}{|c|c|c|c|c|c|c|}
\hline
Number of GPUs & 1 & 2 & 3 & 6 & 9 & 12 \\
\hline
Images per second & 112.2 & 222.6 & 336.8 & 673.7 & 1010.5 & 1383.7 \\
\hline
\end{tabular}
\end{center}
\end{table}

\begin{figure}[h!]
    \centering
    \includegraphics[width=0.8\textwidth]{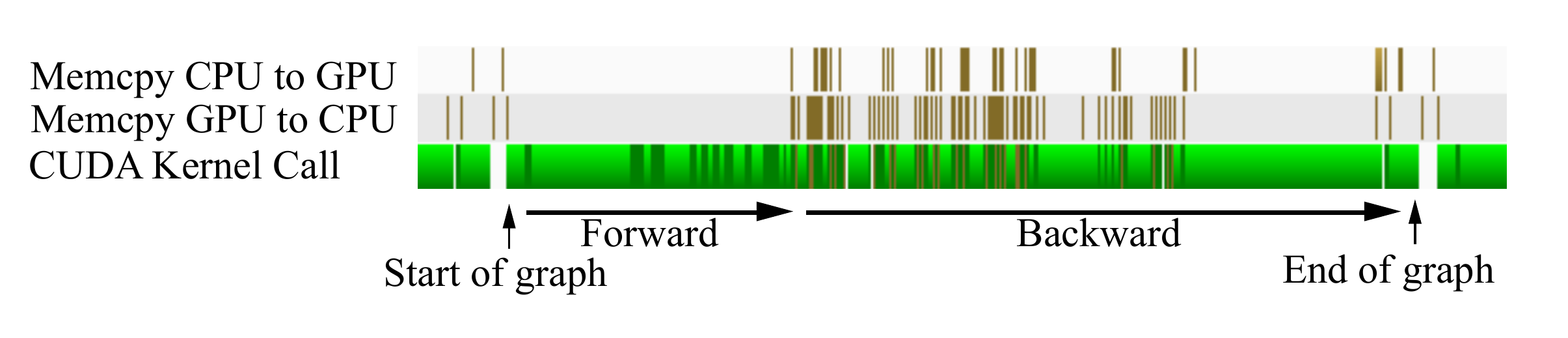}
    \caption{Profiling results of Purine. Memory copies (row 1 and 2)
      are overlapped with computation (row 3). The only overhead is
      the memory copy of first convolution layer, which results in the
      gap between iterations.}
    \label{fig:profile}
\end{figure}

Running GoogLeNet with 4 GPUs on a single machine is profiled and
shown in Figure \ref{fig:profile}. It can be seen that the memory copy
of model parameters between CPUs and GPUs is fully overlapped with the
computations in the backward pass. The only overhead is in the first
layer of the network, which results in the gap between
iterations.

It is favorable to have small batch size in stochastic gradient
descent. However, regarding parallelism, it is more favorable to have
larger batch size and thus higher computation to communication ratio.
We searched for the minumum batch size possible to achieve linear speed
up by exploring different batch sizes.

\begin{table}[h]
\caption{Number of images per second with 12 GPUs and different batch sizes.}
\label{table:batchsizes}
\begin{center}
\begin{tabular}{|c|c|c|c|c|c|c|}
\hline
Batch size per GPU & 128 & 64 & 56 & 48 & 40 & 32 \\
\hline
Images per second & 1383.7 & 1299.1 & 1292.3 & 1279.7 & 1230.2 & 1099.8 \\
\hline
Acceleration Ratio & 12 & 11.26 & 11.20 & 11.09 & 10.66 & 9.53 \\
\hline
\end{tabular}
\end{center}
\end{table}

Table \ref{table:batchsizes} shows the processing speed with different
batch sizes. The acceleration ratio is not reduced much with batch
size 64 as compared to 128. We can still achieve 9.53 fold
acceleration with 12 GPUs when the batch size is set to 32.

\bibliography{iclr2015}

\begin{thebibliography}{12}
\providecommand{\natexlab}[1]{#1}
\providecommand{\url}[1]{\texttt{#1}}
\expandafter\ifx\csname urlstyle\endcsname\relax
  \providecommand{\doi}[1]{doi: #1}\else
  \providecommand{\doi}{doi: \begingroup \urlstyle{rm}\Url}\fi

\bibitem[Bergstra et~al.(2010)Bergstra, Breuleux, Bastien, Lamblin, Pascanu,
  Desjardins, Turian, Warde-Farley, and Bengio]{bergstra+al:2010-scipy}
James Bergstra, Olivier Breuleux, Fr{\'{e}}d{\'{e}}ric Bastien, Pascal Lamblin,
  Razvan Pascanu, Guillaume Desjardins, Joseph Turian, David Warde-Farley, and
  Yoshua Bengio.
\newblock Theano: a {CPU} and {GPU} math expression compiler.
\newblock In \emph{Proceedings of the Python for Scientific Computing
  Conference ({SciPy})}, June 2010.
\newblock Oral Presentation.

\bibitem[Chilimbi et~al.(2014)Chilimbi, Suzue, Apacible, and
  Kalyanaraman]{chilimbi2014project}
Trishul Chilimbi, Yutaka Suzue, Johnson Apacible, and Karthik Kalyanaraman.
\newblock Project adam: Building an efficient and scalable deep learning
  training system.
\newblock In \emph{Proceedings of the 11th USENIX conference on Operating
  Systems Design and Implementation}, pages 571--582. USENIX Association, 2014.

\bibitem[Dean et~al.(2012)Dean, Corrado, Monga, Chen, Devin, Mao, Senior,
  Tucker, Yang, Le, et~al.]{dean2012large}
Jeffrey Dean, Greg Corrado, Rajat Monga, Kai Chen, Matthieu Devin, Mark Mao,
  Andrew Senior, Paul Tucker, Ke~Yang, Quoc~V Le, et~al.
\newblock Large scale distributed deep networks.
\newblock In \emph{Advances in Neural Information Processing Systems}, pages
  1223--1231, 2012.

\bibitem[Isard et~al.(2007)Isard, Budiu, Yu, Birrell, and
  Fetterly]{isard2007dryad}
Michael Isard, Mihai Budiu, Yuan Yu, Andrew Birrell, and Dennis Fetterly.
\newblock Dryad: distributed data-parallel programs from sequential building
  blocks.
\newblock In \emph{ACM SIGOPS Operating Systems Review}, volume~41, pages
  59--72. ACM, 2007.

\bibitem[Jia et~al.(2014)Jia, Shelhamer, Donahue, Karayev, Long, Girshick,
  Guadarrama, and Darrell]{jia2014caffe}
Yangqing Jia, Evan Shelhamer, Jeff Donahue, Sergey Karayev, Jonathan Long, Ross
  Girshick, Sergio Guadarrama, and Trevor Darrell.
\newblock Caffe: Convolutional architecture for fast feature embedding.
\newblock In \emph{Proceedings of the ACM International Conference on
  Multimedia}, pages 675--678. ACM, 2014.

\bibitem[Krizhevsky(2014)]{krizhevsky2014one}
Alex Krizhevsky.
\newblock One weird trick for parallelizing convolutional neural networks.
\newblock \emph{arXiv preprint arXiv:1404.5997}, 2014.

\bibitem[Le(2013)]{le2013building}
Quoc~V Le.
\newblock Building high-level features using large scale unsupervised learning.
\newblock In \emph{Acoustics, Speech and Signal Processing (ICASSP), 2013 IEEE
  International Conference on}, pages 8595--8598. IEEE, 2013.

\bibitem[Low et~al.(2010)Low, Gonzalez, Kyrola, Bickson, Guestrin, and
  Hellerstein]{low2010graphlab}
Yucheng Low, Joseph Gonzalez, Aapo Kyrola, Danny Bickson, Carlos Guestrin, and
  Joseph~M Hellerstein.
\newblock Graphlab: A new framework for parallel machine learning.
\newblock \emph{arXiv preprint arXiv:1006.4990}, 2010.

\bibitem[Olston et~al.(2008)Olston, Reed, Srivastava, Kumar, and
  Tomkins]{olston2008pig}
Christopher Olston, Benjamin Reed, Utkarsh Srivastava, Ravi Kumar, and Andrew
  Tomkins.
\newblock Pig latin: a not-so-foreign language for data processing.
\newblock In \emph{Proceedings of the 2008 ACM SIGMOD international conference
  on Management of data}, pages 1099--1110. ACM, 2008.

\bibitem[Seide et~al.(2014)Seide, Fu, Droppo, Li, and Yu]{seide20141}
Frank Seide, Hao Fu, Jasha Droppo, Gang Li, and Dong Yu.
\newblock 1-bit stochastic gradient descent and its application to
  data-parallel distributed training of speech dnns.
\newblock In \emph{Fifteenth Annual Conference of the International Speech
  Communication Association}, 2014.

\bibitem[Sutskever et~al.(2014)Sutskever, Vinyals, and
  Le]{sutskever2014sequence}
Ilya Sutskever, Oriol Vinyals, and Quoc~VV Le.
\newblock Sequence to sequence learning with neural networks.
\newblock In \emph{Advances in Neural Information Processing Systems}, pages
  3104--3112, 2014.

\bibitem[Szegedy et~al.(2014)Szegedy, Liu, Jia, Sermanet, Reed, Anguelov,
  Erhan, Vanhoucke, and Rabinovich]{szegedy2014going}
Christian Szegedy, Wei Liu, Yangqing Jia, Pierre Sermanet, Scott Reed, Dragomir
  Anguelov, Dumitru Erhan, Vincent Vanhoucke, and Andrew Rabinovich.
\newblock Going deeper with convolutions.
\newblock \emph{arXiv preprint arXiv:1409.4842}, 2014.

\end{thebibliography}
\bibliographystyle{plainnat}

\end{document}